\documentclass[10pt,twocolumn,letterpaper]{article}

\usepackage[pagenumbers]{cvpr}              

\makeatletter
\@namedef{ver@everyshi.sty}{}
\makeatother

\usepackage{graphicx}

\usepackage{xcolor}
\usepackage[pagebackref,breaklinks,colorlinks,bookmarks=false,urlcolor=black,citecolor={green!60!black}]{hyperref}

\usepackage{colortbl}
\definecolor{cGreen}{RGB}{0,255,0}
\definecolor{LightCyan}{rgb}{1,0.88,0.88}

\newcommand{\commentout}[1]{}

\usepackage{tikz}
\usepackage{comment}
\usepackage{amsmath,amssymb} 
\usepackage{color}
\usepackage{times}
\usepackage{epsfig}
\usepackage{amsmath}
\usepackage{amssymb}
\usepackage{booktabs}
\usepackage{caption}
\usepackage{enumitem}
\usepackage{float}
\usepackage[ruled]{algorithm2e}

\usepackage{amsmath,amsfonts,bm}

\def\1{\bm{1}}

\def\vone{{\bm{1}}}

\def\va{{\bm{a}}}

\def\vc{{\bm{c}}}

\def\vs{{\bm{s}}}

\def\vx{{\bm{x}}}
\def\vy{{\bm{y}}}
\def\vz{{\bm{z}}}

\def\mA{{\bm{A}}}

\def\mC{{\bm{C}}}
\def\mD{{\bm{D}}}

\def\mI{{\bm{I}}}

\def\mL{{\bm{L}}}

\def\mW{{\bm{W}}}
\def\mX{{\bm{X}}}
\def\mY{{\bm{Y}}}
\def\mZ{{\bm{Z}}}

\DeclareMathAlphabet{\mathsfit}{\encodingdefault}{\sfdefault}{m}{sl}
\SetMathAlphabet{\mathsfit}{bold}{\encodingdefault}{\sfdefault}{bx}{n}

\newcommand{\tabref}[1]{Table~\ref{tab:#1}}

\DeclareMathOperator*{\argmax}{arg\,max}
\DeclareMathOperator*{\argmin}{arg\,min}

\DeclareMathOperator{\Tr}{Tr}

\usepackage[accsupp]{axessibility}  
\makeatletter
\DeclareRobustCommand\onedot{\futurelet\@let@token\@onedot}
\def\@onedot{\ifx\@let@token.\else.\null\fi\xspace}
\SetKw{kwInit}{Init:}
\SetKw{kwInput}{Input:}
\def\eg{\emph{e.g}\onedot} 
\def\ie{\emph{i.e}\onedot} 
 
 \def\vs{\emph{vs}\onedot}
\def\wrt{w.r.t\onedot} 
\def\etal{\emph{et al}\onedot}

\makeatother

\makeatletter
\def\ps@myheadings{%
    \let\@oddfoot\@empty\let\@evenfoot\@empty
    \def\@evenhead{\thepage\hfil\slshape\leftmark}%
    \def\@oddhead{{\slshape\rightmark}\hfil\thepage}%
    \let\@mkboth\@gobbletwo
    \let\sectionmark\@gobble
    \let\subsectionmark\@gobble
    }
  \if@titlepage
  \renewcommand\maketitle{\begin{titlepage}%
  \let\footnotesize\small
  \let\footnoterule\relax
  \let \footnote \thanks
  \null\vfil
  \vskip 60\p@
  \begin{center}%
    {\LARGE \@title \par}%
    \vskip 3em%
    {\large
     \lineskip .75em%
      \begin{tabular}[t]{c}%
        \@author
      \end{tabular}\par}%
      \vskip 1.5em%
    {\large \@date \par}
  \end{center}\par
  \@thanks
  \vfil\null
  \end{titlepage}%
  \setcounter{footnote}{0}%
}
\else
\renewcommand\maketitle{\par
  \begingroup
    \renewcommand\thefootnote{\@fnsymbol\c@footnote}%
    \def\@makefnmark{\rlap{\@textsuperscript{\normalfont\color{black}\@thefnmark}}}%
    \long\def\@makefntext##1{\parindent 1em\noindent
            \hb@xt@1.8em{%
                \hss\@textsuperscript{\normalfont\@thefnmark}}##1}%
    \if@twocolumn
      \ifnum \col@number=\@ne
        \@maketitle
      \else
        \twocolumn[\@maketitle]%
      \fi
    \else
      \newpage
      \global\@topnum\z@   
      \@maketitle
    \fi
    \thispagestyle{plain}\@thanks
  \endgroup
  \setcounter{footnote}{0}%
}
\makeatother

\usepackage[pagebackref,breaklinks,colorlinks]{hyperref}

\usepackage[capitalize]{cleveref}
\crefname{section}{Sec.}{Secs.}
\Crefname{section}{Section}{Sections}
\Crefname{table}{Table}{Tables}
\crefname{table}{Tab.}{Tabs.}

\def\cvprPaperID{4669} 
\def\confName{CVPR}
\def\confYear{2023}

\usepackage{txfonts}
\newcommand{\heart}{\ensuremath\varheartsuit}

\usepackage{multibib}
\newcites{latex}{References}

\begin{document}

\title{Transductive Few-shot Learning with Prototype-based Label Propagation by Iterative Graph Refinement}

\author{
{Hao Zhu$^{\dagger, \S}$, $\quad$Piotr Koniusz\thanks{The corresponding author.$\quad$This paper is published at CVPR 2023. The code is available at \url{https://github.com/allenhaozhu/protoLP}}$\;^{, \S,\dagger}$}\\
{$^{\dagger}$Australian National University $\quad^\S$Data61\heart CSIRO}\\
allenhaozhu@gmail.com, $\;$firstname.lastname@anu.edu.au\\
}
\maketitle

\begin{abstract}
Few-shot learning (FSL) is popular due to its  ability to adapt to novel classes. 
Compared with inductive few-shot learning, transductive models typically perform better as they leverage all samples of the query set.
The two existing classes of methods, prototype-based and graph-based, have the disadvantages of inaccurate prototype estimation and sub-optimal graph construction with kernel functions, respectively. 
In this paper, we propose a novel prototype-based label propagation to solve these issues.
Specifically, our graph construction is based on the relation between prototypes and samples rather than between samples. As  prototypes are being updated, the graph changes.
%
We also estimate the label of each prototype instead of considering a prototype be the class centre.
On mini-ImageNet, tiered-ImageNet, CIFAR-FS and CUB datasets, we show the proposed method outperforms other state-of-the-art methods in transductive FSL and semi-supervised FSL when  some unlabeled data accompanies the novel few-shot task.
\end{abstract}

\begin{figure}
    \centering
    \includegraphics[width=0.49\textwidth]{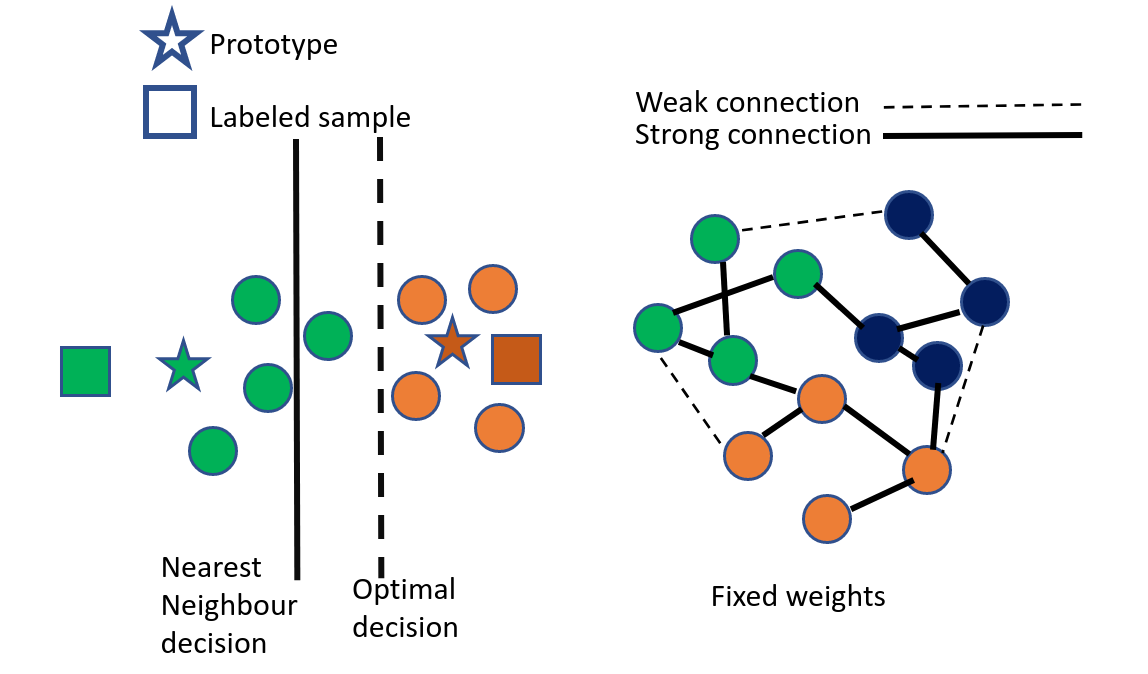}
    \caption{Drawbacks of prototype-based and graph-based FSL. ({\em left}) Some label assignments are incorrect due to the imperfect decision boundary. ({\em right}) Some ``strong'' links in the fixed graph are incorrect as they associate   samples of different classes.}
    \label{fig:issue}
\end{figure}

\section{Introduction}
With the availability of large-scale datasets and the rapid development of deep convolutional architectures, supervised learning exceeds in computer vision, voice, and machine translation~\cite{lecun2015deep}.
However, lack of data makes the existing supervised models fail during the inference on novel tasks. 
As the annotation process may necessitate expert knowledge, annotations are may be scarce and costly (\eg, annotation of medical images). In contrast, humans can learn a novel concept  from just a single example. 

Few-shot learning (FSL) aims to mimic the capabilities of biological vision~\cite{fei2006one}  
%
and it leverages metric learning, meta-learning, or transfer learning. 
The purpose of metric-based FSL is to learn a mapping from images to an embedding space in which images from the same class are closer together and images from other classes are separated. 
Meta-learning FSL performs task-specific optimisation with the goal to generalize to other tasks well. %
Pre-training a feature extractor 
followed by adapting it for reuse 
on new class samples is an example of transfer learning.

Several recent studies~\cite{liu2018learning,hou2019cross,dhillon2019baseline,hu2020empirical,kim2019edge,qiao2019transductive,lazarou2021iterative,qi2021transductive} explored transductive inference for few-shot learning.
At the test time, transductive FSL infer the class label jointly for all the unlabeled query samples, 
rather than for one sample/episode at a time. 
Thus, transductive FSL typically outperforms  inductive FSL. 
%
We categorise transductive FSL into: (i) FSL that requires the use of unlabeled data to estimate prototypes~\cite{snell2017prototypical,simon2020adaptive,lichtenstein2020tafssl,liu2020prototype,boudiaf2020information}, and (ii) FSL that builds a graph with some kernel function and then uses  label propagation 
to predict labels on query sets~\cite{liu2018learning,ziko2020laplacian,lazarou2021iterative}. However, the above two paradigms have their own drawbacks. For prototype-based methods, they usually use the nearest neighbour classifier, which is based on the assumption that there exists an embedding where points cluster around a single prototype representation for each class.
Fig.~\ref{fig:issue} (left) shows a toy example which is sensitive to the large within-class variance and low between-class variance. Thus, the two prototypes cannot be estimated perfectly by the soft label assignment alone. 
Fig.~\ref{fig:issue} (right) shows that Label-Propagation (LP) and Graph Neural Network (GNN) based methods depend on the graph construction which is commonly based on a specific kernel function determining the final result. 
If some  nodes are wrongly and permanently linked, these connections will affect the propagation step.

\begin{figure}
    \centering
    \includegraphics[width=0.49\textwidth]{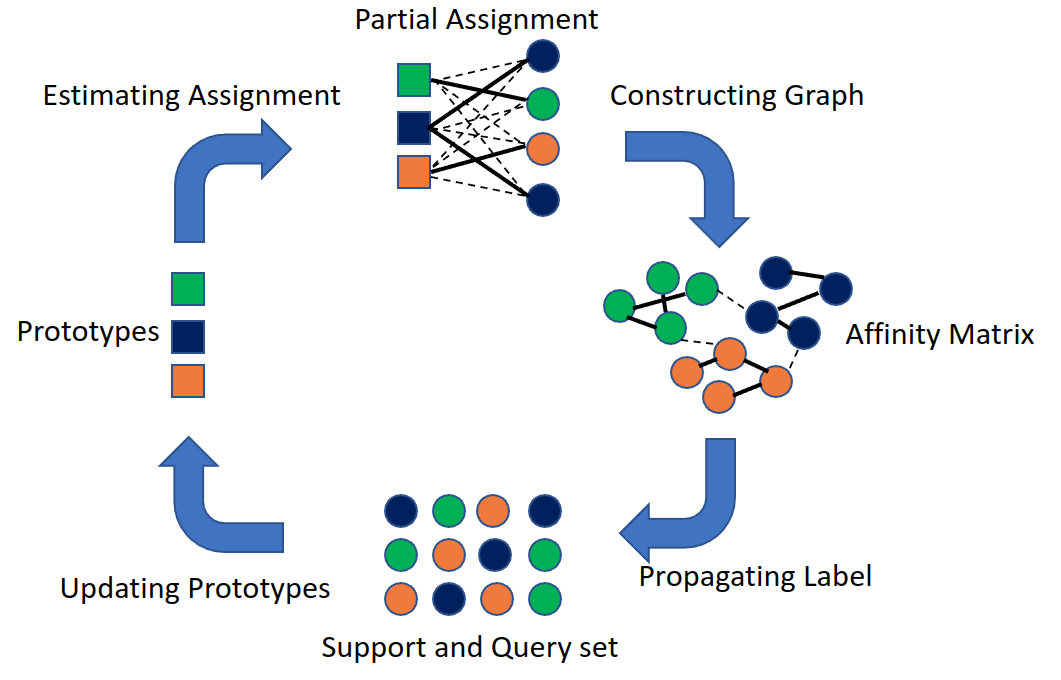}
    \caption{Our   transductive few-shot learning: (i) based on prototypes, estimate as partial assignment (one can use soft k-means in Eq. \eqref{eq:pa} instead); (ii) a graph is constructed by the assignment, followed by the prototype-based label propagation, predicting the label soft score; (ii) updating prototypes based on the prediction.}
    \label{fig:protoLP}
\end{figure}

%
In order to avoid the above pitfalls of transductive FSL, 
we propose {\em prototype-based Label-Propagation} (protoLP). 
Our transductive inference  can work with a generic feature embedding learned on the base classes. 
%
%
%
%
Fig.~\ref{fig:protoLP} shows 
how to alternatively optimize a partial assignment between prototypes and the query set by (i) solving a kernel regression problem (or optimal transport problem) and (ii) a label probability prediction by prototype-based label propagation. 
Importantly, protoLP does not assume the uniform class distribution prior  while significantly outperforming other methods that assume the uniform prior, as shown  
%
in ablations on the imbalanced benchmark~\cite{veilleux2021realistic} 
where methods relying on the balanced class prior fail. 
%
Our model outperforms state-of-the-art methods significantly, consistently providing improvements across different settings, datasets, and training models. 
Our transductive inference is very fast, with runtimes that are close to the runtimes of inductive inference. 

\vspace{0.2cm}
Our contributions are as follows: 
\renewcommand{\labelenumi}{\roman{enumi}.}
\hspace{-1.0cm}
\begin{enumerate}[leftmargin=0.6cm]
\item We identify  issues resulting from  separation of prototype $\!\!$-based and label propagation methods. We propose prototype-based Label Propagation (protoLP) for transductive FSL, which unifies 
both models into one framework. Our protoLP estimates prototypes not only from the partial assignment but also from the prediction of label propagation. The graph for label propagation is not fixed 
as we alternately learn prototypes and the graph.
\item By introducing parameterized label propagation step, we remove the assumption of uniform class prior while other methods highly depend on this prior.
\item  We showcase  advantages of protoLP on four 
datasets for transductive  and semi-supervised learning, Our protoLP outperforms the state of the art 
under various settings including  different backbones, unbalanced query set, and data augmentation.
\end{enumerate}


\section{Related Work}

\noindent\textbf{Few-shot classification}  methods  often exploit the meta-learning paradigm \cite{vinyals2016matching,snell2017prototypical,ravi2017optimization},
and they use episodes for training and testing. 
%
%
Approaches \cite{wang2019simpleshot,chen2019closer} show that meta-training is not required for learning good features for few-shot learning. 
Instead, they train a typical classification network with two blocks: the feature extractor and the classification head.
Many FSL models combine backbone with classification head \cite{maxexp,simon2022wacv,zhang2022accv,wang2022uncertainty,wang2022temporal}, detection head \cite{zhang2020fsod,zhang2022kernelized,zhang2022time}, localization head \cite{lu2022few} or detection head \cite{dahyun2023}.
%
%
%

We focus on 
 designing the inference stage and improving its performance in transductive and semi-supervised setting. 

\vspace{0.1cm}
\noindent\textbf{Graph-based FSL} 
often form a graph via an adjacency matrix based on Radial Basis Function (RBF),  used in the  propagation of labels or features.
Satorras \etal \cite{satorras2018few} propagate labels by building an affinity matrix between the support set and the unlabeled data.
wDAE-GNN~\cite{gidaris2019generating}  generates classification weights with a graph neural network (GNN) and applies a denoising AutoEncoder (DAE) to regularize the representation.
Approach \cite{liu2018learning} learns propagation.  
Embedding Propagation~\cite{rodriguez2020embedding} propagates labels and  the embedding to decrease the intra-class distance. 
Set-to-set functions have also been used for embedding adaptation~\cite{ye2020few}. 

In contrast to 
FSL with a fixed graph, we do not construct a graph from samples directly but construct a bipartite graph by prototypes and samples. As  prototypes change, so does the constructed graph, which we regard as a learnable graph.

\vspace{0.1cm}
\noindent\textbf{Transductive and Semi-Supervised
Few-Shot Learning} is not as popular as  inductive FSL  which only uses samples in the support set  to fine-tune the model or learn a function for the inference of query labels. In contrast,  transductive FSL enjoys access to all the query data. 
In this paper, we categorise transductive FSL into (i) prototype-based and (ii) label propagation based FSL. 
%
Prototypical Network~\cite{snell2017prototypical} learns a metric space in which classification is performed by computing distances to prototype representations of each class. 
Simon \etal \cite{simon2020adaptive} use adaptive subspace-based prototypes. 
Lichtenstein \etal \cite{lichtenstein2020tafssl} employ subspace learning via PCA or ICA to extract discriminant features for the nearest neighbour classifier on prototypes. Liu  \etal \cite{liu2020prototype} improve prototype estimation.  TIM~\cite{boudiaf2020information} maximizes the mutual information between the query features and their label predictions for a few-shot task at inference, while minimizing the cross-entropy loss on the support set to estimate prototypes.
Label Propagation (LP) is popular in transductive FSL  methods~\cite{liu2018learning,chen2021eckpn,ziko2020laplacian,lazarou2021iterative,zhu2022ease}, which construct a graph from the support set and the entire query set, and propagate labels within the graph. However, as in graph-based FSL methods, LP employs kernel functions (RBF or cosine) to construct the graph between samples. 
Additionally, some methods~\cite{hu2020leveraging,lazarou2021iterative,zhu2022ease} leverage the uniform prior on the class distribution with the optimal transport while in realistic evaluation of transductive FSL the prior is unknown~\cite{veilleux2021realistic}.
In semi-supervised FSL~\cite{liu2018learning,li2019learning,ren2018meta,simon2020adaptive}, the unlabeled data is provided in addition to the support set and is assumed to have a similar distribution to the target classes (although unrelated  noisy samples may be also added). 


\section{Methodology}
Below 
%
we introduce prototypical networks~\cite{snell2017prototypical},  explain semi-supervised prototype computation, and present  transductive FSL based on label propagation.
Then, we present our prototype-based Label Propagation (protoLP).
Finally, we show how to optimize protoLP by updating the prototypes, solving the partial assignment problem and the label propagation by the linear projection. Moreover, we demonstrate how to obtain the final label prediction for the query set given learnt prototypes.
Fig.~\ref{fig:protoLP} illustrates our method.

\subsection{Preliminaries}

\noindent\textbf{Inductive FSL} uses a  support set of $K$ classes with $N$ labeled examples per class, $S\equiv\{(\mathbf{x}_i , y_i )\}_{i=1}^L$ where $L=NK$, 
each $\mathbf{x}_i \in \mathbb{R}^D$ is the $D$-dimensional feature vector (from backbone) of an example and $y_i \in \{1,\cdots, K\}$ is the corresponding label.  $S_k\subset S$ is the set of examples labeled with class $k$. Prototypical networks~\cite{snell2017prototypical} compute a prototype of class as the mean vector of support samples belonging to the class:
\begin{equation}
\mathbf{c}_{k}=\frac{1}{\left|S_{k}\right|} \sum_{\mathbf{x}:\, y(x)\in S_{k}} \mathbf{x}.
\label{eq:pc}
\end{equation}

Given a distance function $d : \mathbb{R}^D \times \mathbb{R}^D \rightarrow \mathbb{R}^+$, prototypical nets use Nearest Class Mean (NCM) to predict the label of  query  $\mathbf{x}$: 
\begin{equation}
k^*    =\arg\min_k d\left(\mathbf{x}, \mathbf{c}_{k}\right).
\label{eq:pn}
\end{equation}

\vspace{0.1cm}
\noindent\textbf{Transductive Few-shot Learning.} In the case of inductive FSL, the prediction is performed independently on each episode, and thus the mean vector is only dependent on the support set of $N$ labeled examples, as shown in Eq.~\eqref{eq:pc}, and is fixed for the given embedded features. However, in the case of transductive FSL, the prediction is performed inclusive of all queries, $Q \equiv \{\mathbf{x}_{L+i}\}^U_{i=1}$, where $U=RK$, and the query set has $K$ classes with $R$ unlabeled examples per class. 

\noindent\textbf{Inference of Prototypes.} Transductive/semi-supervised Prototypical Network~\cite{snell2017prototypical} treats prototypes $\vc_k$ in Eq.~\eqref{eq:pc} as clusters. The unlabeled samples with indexes $L+1\leq i\leq L+U$ are soft-assigned \cite{pk_sa}  to each cluster $\mathbf{c}_k$, yielding $z_{ik}$, whereas labeled samples with indexes $1\leq i\leq L$ use one-hot labels, \ie, $z_{ik}=1\text{ for }k=y_i$ and $z_{ik}=0\text{ for }k\neq y_i$. 
Specifically, refined prototypes are obtained 
as follows:$\!\!\!$
\begin{align}
\vc_{k}&=\frac{\sum_{i=1}^L z_{ik}\vx_{i}+\sum_{j=L+1}^{L+U} {z}_{jk}\vx_{j}}{\sum_{i'=1}^L z_{i' k}+\sum_{j'=L+1}^{L+U} {z}_{j'k}}\qquad\text{where}\\
{z}_{ik}&=\begin{cases}\frac{\exp \left(-\lVert\vx_i-\vc_{k}\rVert_{2}^{2}\right)}{\sum_{k^{\prime}} \exp \left(-\lVert\vx_{i}-\boldsymbol{c}_{k^\prime}\rVert_{2}^{2}\right)} \quad\text{if}\quad L+1\leq i\leq L+U\\
\text{OneHot}(y_i) \quad\qquad\text{if}\quad 1\leq i \leq L.\end{cases}
\label{eq:pa}
\end{align}
The prediction of each query 
label follows Eq.~\eqref{eq:pn}. 
Notice that although the prototypes estimation leverages all data in the query set, the inference still only depends on prototypes and a single sample rather than prototypes and all samples.

\vspace{0.1cm}
\noindent\textbf{Label Propagation.} We form a graph $G = (\mathcal{V},\mathcal{E})$ where  vertices $\mathcal{V}$ represent all labeled and unlabeled samples, and  edges $\mathcal{E}$ are represented by a distance matrix $\mW$. Let $\mD$ be a diagonal matrix whose diagonal elements are given by $D_{ii} = \sum_j{W_{ij}}$. The graph Laplacian is then defined as $\mL = \mD-\mW$, which is used for smoothness-based regularization by taking into account the unlabeled data:
\begin{equation}
\frac{1}{2}\Tr(\tilde{\mY}^{\top}\mL\tilde{\mY})= \frac{1}{2}\sum_{i,j} W_{ij}(\tilde{\vy_i}-\tilde{\vy_j})^2.
\end{equation} 
%
For practical reasons, 
Zhou \etal \cite{zhou2003learning} are concerned not only with the smoothness but  the impact of the supervised loss on the propagation. Thus, they minimize a combination of the smoothness and the squared error training loss:
\begin{equation}
 \tilde{\mY}^* = \argmin_{\tilde{\mY}}\sum_{i=1}^L \|\tilde{\vy}_i-\vy_i\|_2^2+\frac{\lambda}{2}\Tr( \tilde{\mY}^{\top}\mL\tilde{\mY}).
\label{eq:ssl}
\end{equation}

Eq. \eqref{eq:ssl} relies on the quality of a fixed Laplacian matrix $\mL$ which largely determines the final performance of method. 

\subsection{Proposed Formulation}
Below we introduce our prototype-based Label Propagation (protoLP). Firstly, we  parameterize the label propagation step and explain why. 
Then, we explain how to use prototypes to construct a graph,  and we   combine the above two components into protoLP.

\vspace{0.1cm}
\noindent{\textbf{Parameterized Label Prediction}.} Given the adjacency matrix $\mW$, we can solve  label propagation by Eq.~\eqref{eq:ssl}. However, we 
introduce a linear projection $\mA$ into the label propagation step to limit overfitting to matrix  $\mW$. Let:
\begin{equation}
\tilde{\mY} = \mZ\mA,
\label{eq:lr}
\end{equation}
%
\noindent
where $\mA = [\va_1,\cdots,\va_K]^\top$ has $K$ basis functions and $\mZ$ comes from Eq.~\eqref{eq:pa} given a prototype set $\{\vc_k\}_{k=1}^K$. Substitute Eq. \eqref{eq:lr} into Eq. \eqref{eq:ssl}, we obtain:
\begin{equation}
\!\!\!\!\mA^*\!=\mathop{\arg \min}_{\mA}\left \|\mZ_L\mA -\mY_L\right \|^2_F + \frac{\lambda}{2}\Tr(\mA^\top\mZ^\top\mL\mZ\mA),
\label{eq:AGSSL}
\end{equation}
%
\noindent
where $\mZ_L \!=\! [\vz_1,\dots,\vz_L]^\top\!\in\! \mathbb{R}^{L\times K}$. $\mY_L\! =\! [\vy_1,\dots,\vy_L]^\top\!\in\! \mathbb{R}^{L\times K}$ is the submatrix according to the assignment and label partition. Intuitively, we can regard $\va_k$ as a learnable label for the $k$-th prototype which is non-sparse in contrast to a one-hot class vector. Based on the above model, 
one can estimate a  soft score  of likely category of an inaccurate prototype. 

\vspace{0.1cm}
\noindent{\textbf{Prototype-based Graph Construction}.} Prototype-based graphs are based on the idea that we can use a small number of prototypes to turn sample-to-sample affinity computations into much simpler sample-to-prototype affinity computations~\cite{liu2010large}. Below we explain how to construct a graph with prototypes. Given a prototype set $\{\vc_k\}_{k=1}^K$, for each sample we  obtain a partial assignment $\vz_j$ with soft assignment in Eq.~\eqref{eq:pa}. We form the adjacency matrix $\mW$ as: 
\begin{equation}
\mW=\mZ \mathbf{\Lambda}^{-1} \mZ^{\top},
\label{eq:affmat}
\end{equation}
%
%
\noindent
where the diagonal matrix $\mathbf{\Lambda} \in \mathbb{R}^{K \times K}$ is defined as $\Lambda_{k k}=\sum_{i} Z_{i k}$ (index $i$ iterates over all samples). The corresponding Laplacian matrix is $\mL=\mI-\mZ \mathbf{\Lambda}^{-1} \mZ^{\top}$.  $W_{ij}$ captures relation between the $i$-th  and  $j$-th samples 
by confounding variables $\mathbf{c}_k$ according to the chain rule of Markov random walks:
\vspace{-0.2cm}
\begin{equation}
\begin{aligned}
W_{ij} \!=\! p\left(\vx_{i}\!\mid\!\vx_{j}\right) &=\sum_{k=1}^{K} p\left(\vx_{i}\!\mid \!\vc_{k}\right) p\left(\vc_{k}\!\mid\!\vx_{j}\right)\!=\!p\left(\vx_{j}\!\mid\!\vx_{i}\right) \\
&=\sum_{k=1}^{K} \frac{z_{jk}}{\sum_{j^{\prime}} z_{j^{\prime}k}} z_{i,k}=\sum_{k=1}^{K} \frac{z_{ik} z_{jk}}{\Lambda_{k k}},
\end{aligned}
\end{equation}
%
%
\noindent
where $p(\vx_i\!\mid\!\vc_k)\!=\!Z_{ik}$  
and $W_{ij}\!=\!W_{ji}$. One may think of the above process as a 2-hop diffusion on a bipartite graph with samples $\mathbf{x}_i$ and prototypes $\mathbf{c}_k$ located in two partitions of that graph. Notice  the graph changes with prototypes. 

\vspace{0.1cm}
\noindent\textbf{Prototype-based Label Propagation (protoLP).} Based on parameterized label prediction and prototype-based graph construction, we  combine Eq.~\eqref{eq:AGSSL} and \eqref{eq:affmat} into:
\begin{equation}
\!\!\!\!\min_{\mA}\frac{1}{2}\left \|\mZ_L\mA -\mY_L\right \|^2_F + \frac{\lambda}{2}\Tr(\mA^\top\mZ^\top(\mI - \mZ\mathbf{\Lambda}^{-1}\mZ^\top)\mZ\mA).
\label{eq:protoLP}
\end{equation}
As Eq.~\eqref{eq:AGSSL} and \eqref{eq:affmat} are highly dependent on prototypes, instead of using the update of $\mathbf{c}_k$ as in Eq.~\eqref{eq:pa}, we use steps from Section \ref{sec:opt}. 
Firstly, we initialize each prototype as the mean vector of the support samples belonging to its class. 

\begin{algorithm}[t]
\caption{Prototype-based Label Propagation.}
\kwInput{$\mX, \mY, \lambda, \alpha, n_\text{step}$}\\
\kwInit{$\mathbf{\tilde{c}}_{k}=\frac{1}{\left|S_{k}\right|} \sum_{\left(\vx_i,y_{i}\right) \in S_{k}} \vx_i$,$k=0$\;}
 \While{$k < n_\text{step}$}{
 \vspace{0.1cm}
 {\em Estimating Assignment:}\\
 $\;\;{Z}_{ij}=\frac{\exp(-\left\|\vx_i-\tilde{\vc}_j\right\|^2_2)}{\sum_{j'}exp(-\left\|\vx_i-\tilde{\vc}_{j'}\right\|^2_2)}$\;
 \vspace{0.1cm}
{\em Constructing Graph:}\\
 $\;\;\Lambda_{kk} = \sum_iZ_{ik}$ and $\mW=\boldsymbol{Z}_t \boldsymbol{\Lambda}^{-1} \boldsymbol{Z}^{\top}_t$\;
 \vspace{0.1cm}
 {\em Propagating Label:}\\
 $\;\;\tilde{\mathbf{Y}}=\boldsymbol{Z}_t\left(\boldsymbol{Z}_L^{\top} \boldsymbol{Z}_L+\lambda \boldsymbol{Z}^{\top}_t\left(\boldsymbol{I}-\mW\right) \boldsymbol{Z}_t\right)^{-1} \boldsymbol{Z}^{\top} _t\boldsymbol{Y}$\;
 \vspace{0.1cm}
{\em Updating Prototypes:}\\
  $\;\;\mathbf{\tilde{C}} \leftarrow (1-\alpha)\mathbf{\tilde{C}}+\alpha\Tilde{\mY}\mX$\;
$k\leftarrow k + 1$
 }
\Return $y_{i}=\argmax_j {\Tilde{Y}}_{i,j}$
\label{alg:protoLP}
\end{algorithm}

\begin{table*}[htbp]
  \centering
  \caption{Comparison of test accuracy against state-of-the-art methods for 1-shot and 5-shot classification. ($^*$: inference aug., \S \ref{sec:da})
  }
  \resizebox{1\textwidth}{!}{
    \begin{tabular}{l c c ll cc}
    \toprule
          &      &       & \multicolumn{2}{c}{mini-ImageNet} & \multicolumn{2}{c}{tiered-ImageNet} \\
    \midrule
    Methods & Setting &Network & 1-shot &  5-shot & 1-shot &  5-shot \\
    \midrule
    MAML~\cite{finn2017model} & Inductive & ResNet-18 & 49.61 $\pm$ 0.92 & 65.72 $\pm$ 0.77 & --   & --  \\
    RelationNet~\cite{sung2018learning} & Inductive & ResNet-18 & 52.48 $\pm$ 0.86 &  69.83 $\pm$ 0.68 & --   & --  \\
    MatchingNet~\cite{vinyals2016matching} & Inductive & ResNet-18 & 52.91 $\pm$ 0.88 &  68.88 $\pm$ 0.69 & --   & --  \\
    ProtoNet~\cite{snell2017prototypical} & Inductive & ResNet-18 & 54.16 $\pm$ 0.82  & 73.68 $\pm$ 0.65 & --   & -- \\
    TPN~\cite{liu2018learning} & transductive & ResNet-12 & 59.46 & 75.64 & --   & --\\
    TEAM~\cite{qiao2019transductive} & transductive & ResNet-18 & 60.07 & 75.9  & --   & --  \\
    Transductive tuning~\cite{dhillon2019baseline} & Transductive & ResNet-12 & 62.35 $\pm$ 0.66 &  74.53 $\pm$ 0.54 & --   & -- \\
    MetaoptNet~\cite{lee2019meta} & Transductive & ResNet-12 & 62.64 $\pm$ 0.61 &  78.63 $\pm$ 0.46 & 65.99 $\pm$ 0.72  & 81.56 $\pm$ 0.53 \\
    CAN+T~\cite{hou2019cross} & Transductive & ResNet-12 & 67.19 $\pm$ 0.55 &  80.64 $\pm$ 0.35 & 73.21 $\pm$ 0.58  & 84.93 $\pm$ 0.38 \\
    DSN-MR~\cite{simon2020adaptive} & Transductive & ResNet-12 & 64.60 $\pm$ 0.72 &	79.51 $\pm$ 0.50 & 67.39 $\pm$ 0.82 &	82.85 $\pm$ 0.56 \\
    ODC$^*$~\cite{qi2021transductive} &Transductive & ResNet-18 & 77.20 $\pm$ 0.36 & 87.11 $\pm$ 0.42 & 83.73 $\pm$ 0.36 & 90.46 $\pm$ 0.46 \\
    MCT$^*$~\cite{kye2020meta} &Transductive & ResNet-12 & 78.55 $\pm$ 0.86 & 86.03 $\pm$ 0.42 &82.32 $\pm$ 0.81 & 87.36 $\pm$ 0.50 \\
    EASY$^*$\cite{bendou2022easy} &Transductive & ResNet-12 & 82.31 $\pm$ 0.24 & 88.57 $\pm$ 0.12 & 83.98 $\pm$ 0.24 & 89.26 $\pm$ 0.14 \\
    \rowcolor{LightCyan}  protoLP (ours) & Transductive & ResNet-12 & \textcolor{red}{70.77 $\pm$ 0.30}  &   \textcolor{red}{80.85 $\pm$ 0.16} & \textcolor{red}{84.69 $\pm$ 0.29}  & \textcolor{red}{89.47 $\pm$ 0.15} \\
    \rowcolor{LightCyan}  protoLP$^*$ (ours) & Transductive & ResNet-12 & \textcolor{red}{84.35 $\pm$ 0.24}  &   \textcolor{red}{90.22 $\pm$ 0.11} & \textcolor{red}{86.27 $\pm$ 0.25}  & \textcolor{red}{91.19 $\pm$ 0.14} \\    
    \rowcolor{LightCyan}  protoLP (ours) & Transductive & ResNet-18 & \textcolor{red}{75.77 $\pm$ 0.29}  &   \textcolor{red}{84.00 $\pm$ 0.16} & \textcolor{red}{82.32 $\pm$ 0.27}  & \textcolor{red}{88.09 $\pm$ 0.15} \\
    \rowcolor{LightCyan}  protoLP$^*$ (ours) & Transductive & ResNet-18 & \textcolor{red}{85.13 $\pm$ 0.24}  &   \textcolor{red}{90.45 $\pm$ 0.11} & \textcolor{red}{83.05 $\pm$ 0.25}  & \textcolor{red}{88.62 $\pm$ 0.14} \\
    \midrule
    ProtoNet~\cite{snell2017prototypical} & Inductive & WRN-28-10   & 62.60 $\pm$ 0.20 &  79.97 $\pm$ 0.14 & --   & --  \\
    MatchingNet~\cite{vinyals2016matching} & Inductive & WRN-28-10   & 64.03 $\pm$ 0.20  & 76.32 $\pm$ 0.16 & --   & -- \\
    SimpleShot~\cite{wang2019simpleshot} & Inductive & WRN-28-10   & 65.87 $\pm$ 0.20 &  82.09 $\pm$ 0.14 & 70.90 $\pm$ 0.22 &  85.76 $\pm$ 0.15 \\
    S2M2-R~\cite{mangla2020charting} & Inductive & WRN-28-10   & 64.93 $\pm$ 0.18 &	83.18 $\pm$ 0.11 & - &  - \\
    Transductive tuning~\cite{dhillon2019baseline} & Transductive & WRN-28-10   & 65.73 $\pm$ 0.68  & 78.40 $\pm$ 0.52 & 73.34 $\pm$ 0.71 &  85.50 $\pm$ 0.50 \\
    SIB~\cite{hu2020empirical} & Transductive & WRN-28-10   & 70.00 $\pm$ 0.60  & 79.20 $\pm$ 0.40 & --   &--  \\
    BD-CSPN~\cite{liu2020prototype} & Transductive & WRN-28-10   & 70.31 $\pm$ 0.93  & 81.89 $\pm$ 0.60 & 78.74 $\pm$ 0.95 &  86.92 $\pm$ 0.63 \\
    EPNet~\cite{rodriguez2020embedding} & Transductive & WRN-28-10 & 70.74 $\pm$ 0.85 & 84.34 $\pm$ 0.53 & 78.50 $\pm$ 0.91 & 88.36 $\pm$ 0.57 \\
    LaplacianShot~\cite{ziko2020laplacian} & Transductive & WRN-28-10   & 74.86 $\pm$ 0.19 &  84.13 $\pm$ 0.14 & 80.18 $\pm$ 0.21 &  87.56 $\pm$ 0.15 \\
    
    ODC~\cite{qi2021transductive} & Transductive & WRN-28-10 & 80.22  &   {88.22} & {84.70}  & {91.20} \\
    iLPC~\cite{lazarou2021iterative} & Transductive & WRN-28-10 & 83.05 $\pm$ 0.79 & 88.82 $\pm$ 0.42 &88.50 $\pm$ 0.75 &92.46 $\pm$ 0.42 \\
    \rowcolor{LightCyan} protoLP (ours) & Transductive & WRN-28-10   & \textcolor{red}{83.07 $\pm$ 0.25} &  \textcolor{red}{89.04 $\pm$ 0.13} & \textcolor{red}{89.04 $\pm$ 0.23} &  \textcolor{red}{92.80 $\pm$ 0.13} \\
    \rowcolor{LightCyan} protoLP* (ours) & Transductive & WRN-28-10   & \textcolor{red}{84.32 $\pm$ 0.21} &  \textcolor{red}{90.02 $\pm$ 0.12} & \textcolor{red}{89.65 $\pm$ 0.22} &  \textcolor{red}{93.21 $\pm$ 0.13} \\
    \bottomrule
    \end{tabular}
    }
    \label{tab:result}
\end{table*}

\subsection{Optimization}
\label{sec:opt}
Below we explain how to optimize \wrt $\mZ,\mA$ and $\mC$ by alternating. The order of optimisation in each round assumes minimization \wrt $\mZ$, then $\mA$ and finally $\mC$. 

\vspace{0.1cm}
\noindent\textbf{Updating} $\mathbf{Z}$. Firstly, given a prototype set $\{\vc_k\}_{k=1}^K$, we  optimize the following equation \wrt $\mathbf{Z}$:
\begin{equation}
    \mathbf{Z}_{t} = \argmin_{\mZ} \sum_{i,k} z_{ik} \|\vx_i-\vc_k\|^2_2,\;\text{s.t.}\;\sum_k\vz_{ik}\!=\!1.
\label{eq:soft}
\end{equation}
%
\noindent
Eq.~\eqref{eq:soft} can be solved by Eq.~\eqref{eq:pa}.
\vspace{0.1cm}
\noindent\textbf{Updating} $\mathbf{A}$. Next, we solve Eq.~\eqref{eq:protoLP} \wrt  $\mathbf{A}$ by globally-optimal  closed-form formula:
\begin{equation}
\!\!\!\!\mA_t\!=\!\left(\mZ_L^{\top}\mZ_L+\lambda \mZ_t^\top\!\left(\mI - \mZ_t\mathbf{\Lambda}^{-1}\mZ_t^\top\right)\mZ_t\right)^{-1}\!\mZ_t^{\top}\mY.
\label{eq:solveA}
\end{equation} 
Subsequently, we can infer the label soft score by Eq.~\eqref{eq:lr}, \ie,
 $\tilde{\mY}_t=\mZ_t\mA_t$ is the output for updating prototypes in the next iteration. Substituting Eq.~\eqref{eq:solveA} into \eqref{eq:lr}, we  have:
\begin{equation}
    \mathbf{\tilde{Y}}_t = \mZ_t(\mZ_L^{\top}\mZ_L+\lambda  \mZ_t^\top\left(\mI - \mZ_t\mathbf{\Lambda}^{-1}\mZ_t^\top\right)\mZ_t)^{-1}\mZ_t^{\top}\mY.
\end{equation}
Using $\mA$ is not mandatory but this linear projection  improves results by limiting overfitting during propagation. 


\vspace{0.1cm}
\noindent\textbf{Updating $\mathbf{C}$.} We update $\mathbf{C}$ by:
\begin{align}
\mathbf{C}_t &= \argmin_{\mathbf{C}} \sum_{i,k} \tilde{y}_{ik} \|\vx_i-\vc_k\|^2_2,
\\
&\!\!\!\!\!\text{where one may set }\;\vc_{k}=\frac{\sum_{i=1}^L \vx_{i} \tilde{y}_{ik}+\sum_{j=L+1}^{L+U} \vx_{j}\tilde{y}_{j k}}{\sum_{i^\prime=1}^L \tilde{y}_{i^\prime k}+\sum_{j^\prime=L+1}^{L+U} \tilde{y}_{j^\prime k}}.
\end{align}
We use the gradient decent  and the exponential running average  to update $\mC$
to avoid instability that changing $\tilde{\mY}$ may pose:
\begin{equation}
\mathbf{C}_t = (1-\alpha)\mathbf{C}_{t-1} + \alpha \tilde{\mY}^\top\!\mX, 
\label{eq:abc}
\end{equation}
where $0\leq\alpha\leq 1$  controls the speed of adaptation of $\mathbf{C}_t$.

\vspace{0.1cm}
\noindent\textbf{Inference.} For each query $\vx_{L+i}\!\in\!Q$ (note $\vx_1,\cdots,\vx_L$ where $L\!=\!NK$ are for training, we predict its pseudo-label by $\argmax\limits_{k\in\{1,\cdots,K\}} \tilde{y}_{jk}$ that corresponds to the maximum element of the $j$-th row of the resulting matrix $\tilde{\mY}$.

\vspace{0.1cm}
\noindent\textbf{Uniform Prior with Optimal Transport.}
In transductive FSL, the evaluation based on the balanced class setting is widely used. Thus many methods~\cite{lazarou2021iterative,hu2020leveraging,zhu2022ease} leverage the prior of uniform distribution of class labels to improve the performance. We also consider this factor and normalize $\tilde{\mY}$ to a given row-wise sum $\mathbf{d}_r \in \mathbb{R}^{U}$ and column-wise sum $\mathbf{d}_c \in \mathbb{R}^{K}$. The normalization itself is a projection of $\tilde{\mY}$ onto the set $\mathbb{S}_{\mathbf{d}_r, \mathbf{d}_c}$ of non-negative $U \times K$ matrices having row-wise sum $\mathbf{d}_r$ and column-wise sum $\mathbf{d}_c$:$\!\!\!$
\begin{equation}
\mathbb{S}_{\mathbf{d}_r, \mathbf{d}_c}\!\equiv\!\left\{\tilde{\mY}\!\in\!\mathbb{R}^{U \times K}\!: \tilde{\mY} \mathbf{1}_{K}\!=\!\mathbf{d}_r, \tilde{\mY}^{\top}\! \mathbf{1}_{U}\!=\!\mathbf{d}_c\right\}.
\label{eq:sinkh}
\end{equation}
We use the Sinkhorn-Knopp algorithm~\cite{knight2008sinkhorn} for this projection. It alternates (until convergence) between rescaling the rows of $\tilde{\mY}$ to add up to $\mathbf{d}_r$ and its columns to add up to $\mathbf{d}_c$:
$$
\begin{array}{l}
\tilde{\mY} \leftarrow \operatorname{diag}(\mathbf{d}_r) \operatorname{diag}\left(\tilde{\mY} \mathbf{1}_{U}\right)^{-1} \tilde{\mY}, \\
\tilde{\mY} \leftarrow \tilde{\mY} \operatorname{diag}\left(\tilde{\mY}^{\top} \mathbf{1}_{K}\right)^{-1} \operatorname{diag}(\mathbf{d}_c).
\end{array}
$$
For the uniform prior assumption, $\mathbf{d}_c = \vone_U$ and $\mathbf{d}_c = R\cdot\vone_{K}$, where $R$ is the query number of each class.

\vspace{0.1cm}
\noindent\textbf{Algorithm \ref{alg:protoLP}} $\!\!$summarizes our standard protoLP  without the Sinkhorn-Knopp algorithm~\cite{cuturi2013sinkhorn} omitted for brevity. 
Four steps indicated in italics  are also indicated in Fig. \ref{fig:protoLP}.

\section{Experiments}
We evaluate our method on four few-shot classification benchmarks, mini-ImageNet~\cite{vinyals2016matching}, tiered-ImageNet~\cite{ren2018meta}, CUB~\cite{welinder2010caltech} and CIFAR-FS~\cite{chen2019closer,krizhevsky2009learning}, all often used in  transductive and semi-supervised FSL~\cite{kim2019edge,liu2018learning,qiao2019transductive,ren2018meta,simon2020adaptive}.
%
We use the standard evaluation protocols. 
The results of the transductive and semi-supervised FSL evaluation together with comparisons to previous methods are summarized in Tables~\ref{tab:result},~\ref{tab:cub},~\ref{tab:cifar}, \ref{tab:result2} and \ref{tab:result3}, and  discussed below. 
The performance numbers are given as accuracy \%, and the 0.95 confidence intervals are reported. 
The tests are performed on 10,000 randomly drawn 5-way episodes, with 1 or 5 shots (number of support examples per class), and with 15 queries per episode (unless otherwise specified).
We use publicly available pre-trained backbones 
that are trained on the base class training set. We experiment with 
 ResNet-12, ResNet-18~\cite{oreshkin2018tadam}, and WRN-28-10~\cite{rusu2018meta} backbones pre-trained in S2M2-R~\cite{mangla2020charting}, and DenseNet~\cite{lichtenstein2020tafssl} and  MobileNet~\cite{howard2017mobilenets} pre-trained in SimpleShot~\cite{wang2019simpleshot}. 


\subsection{FSL benchmarks used in our experiments}

\noindent\textbf{Transductive FSL Setting.}
We investigate  transductive FSL  with the set of queries  as the source of  unlabeled data, which is typical  when an FSL classifier receives a bulk of the query data for an off-line evaluation. In \tabref{result}, we report the performance of our protoLP, and compare it to baselines and state-of-the-art (SOTA) transductive FSL methods from the literature: TPN~\cite{liu2018learning}, Transductive FineTuning~\cite{dhillon2019baseline}, MetaOptNet~\cite{lee2019meta}, DSN-MR~\cite{simon2020adaptive}, EPNet~\cite{rodriguez2020embedding}, CAN-T~\cite{hou2019cross}, SIB~\cite{xu2021attentional}, BP-CSPN~\cite{liu2018learning}, LaplacianShot~\cite{ziko2020laplacian}, RAP-LaplacianShot~\cite{hong2021reinforced}, ICI~\cite{wang2020instance}, TIM~\cite{boudiaf2020information}, iLPC~\cite{lazarou2021iterative}, and PT-MAP~\cite{hu2020leveraging}. We also compare to SOTA regular FSL based on S2M2-R~\cite{mangla2020charting} to highlight the effectiveness of using the unlabeled data. 
Tables \ref{tab:result}, \ref{tab:cub} and \ref{tab:cifar} show that on both transductive FSL benchmarks (mini-ImageNet and tiered-ImageNet), protoLP consistently outperforms all the previous (transductive and inductive) methods. 

Our protoLP is insensitive to the feature extractor, \eg, see protoLP with DenseNet and MobileNet  in Table~\ref{tab:result3}. We used 
 features  from SimpleShot~\cite{wang2019simpleshot} backbones. 
%
Compared with SimpleShot, protoLP  gains 13\% and 3\% on the 1-  and  5-shot protocols. 
It also outperforms other transductive methods based on DenseNet such as LaplacianShot~\cite{ziko2020laplacian}, RAP-LaplacianShot~\cite{hong2021reinforced} and variants of TAFSSL~\cite{lichtenstein2020tafssl}.

\begin{table}
\begin{minipage}[b]{0.48\textwidth}
\centering
\caption{Test accuracy \vs the state of the art (transductive inference, 1- and 5-shot classification, CUB). ($^*$: inference aug., \S \ref{sec:da})}
\vspace{-0.1cm}
\resizebox{1\textwidth}{!}{
\begin{tabular}{l l l l}
\toprule
\multicolumn{4}{c}{CUB}                                               \\
\midrule
Method                     & Backbone & 1-shot        & 5-shot        \\
\midrule
LaplacianShot~\cite{ziko2020laplacian} & ResNet-18 &80.96 & 88.68 \\
LR+ICI~\cite{wang2020instance} & ResNet-12 & 86.53$\pm$0.79 & 92.11$\pm$0.35 \\
iLPC~\cite{lazarou2021iterative}     & ResNet-12 & 89.00$\pm$0.70 & 92.74$\pm$0.35 \\
\rowcolor{LightCyan} protoLP  (ours)               & ResNet-12      & \textcolor{red}{90.13$\pm$0.20} & \textcolor{red}{92.85$\pm$0.11} \\
\rowcolor{LightCyan} protoLP$^*$ (ours) & ResNet-12 & \textcolor{red}{91.82$\pm$0.18} & \textcolor{red}{94.65$\pm$0.10} \\
\midrule
BD-CSPN~\cite{liu2020prototype}         & WRN-28-10      & 87.45       & 91.74       \\
TIM-GD~\cite{boudiaf2020information}     & WRN-28-10      & 88.35$\pm$0.19 & 92.14$\pm$0.10 \\
PT+MAP~\cite{hu2020leveraging} & WRN-28-10 & 91.37$\pm$0.61 & 93.93$\pm$0.32 \\
LR+ICI~\cite{wang2020instance} & WRN-28-10 & 90.18$\pm$0.65 & 93.35$\pm$0.30 \\
iLPC~\cite{lazarou2021iterative}     & WRN-28-10 & 91.03$\pm$0.63 & 94.11$\pm$0.30 \\
\rowcolor{LightCyan} protoLP (ours)               & WRN-28-10      & \textcolor{red}{91.69$\pm$0.18} & \textcolor{red}{94.18$\pm$0.09}\\
\bottomrule
\end{tabular}
}
\label{tab:cub}
\end{minipage}

\vspace{0.3cm}
\begin{minipage}[b]{0.48\textwidth}
\centering
\caption{Test accuracy \vs state of the art (transductive inference, 1- and 5-shot classification, CIFAR-FS). ($^*$: inference aug., \S \ref{sec:da})}
\vspace{-0.1cm}
\resizebox{1\textwidth}{!}{
\begin{tabular}{l l l l}
\toprule
\multicolumn{4}{c}{CIFAR-FS}                                               \\
\midrule
Method                     & Backbone & 1-shot        & 5-shot        \\
\midrule
LR+ICI~\cite{wang2020instance} & ResNet-12 & 75.36$\pm$0.97 & 84.57$\pm$0.57  \\
iLPC~\cite{lazarou2021iterative}      & ResNet-12 & 77.14$\pm$0.95 & 85.23$\pm$0.55  \\
DSN-MR~\cite{simon2020adaptive} &	ResNet-12 &	75.60$\pm$0.90	& 85.10$\pm$0.60 \\
SSR~\cite{shen2021re} & ResNet-12	& 76.80$\pm$0.60	& 83.70$\pm$0.40 \\
\rowcolor{LightCyan} protoLP (ours)               & ResNet-12      & \textcolor{red}{78.66$\pm$0.24} & \textcolor{red}{85.85$\pm$0.17}\\
\rowcolor{LightCyan} protoLP$^*$ (ours)  & ResNet-12 & \textcolor{red}{88.22 $\pm$0.21} & \textcolor{red}{91.52$\pm$0.15}\\
\midrule
SIB~\cite{hu2020empirical} &WRN-28-10 &	80.00$\pm$0.60	&85.30$\pm$0.40 \\
PT+MAP~\cite{hu2020leveraging} & WRN-28-10  & 86.91$\pm$0.72 & 90.50$\pm$0.49  \\
LR+ICI~\cite{wang2020instance} & WRN-28-10  & 84.88$\pm$0.79 & 89.75$\pm$0.48  \\
iLPC~\cite{lazarou2021iterative}      & WRN-28-10  & 86.51$\pm$0.75 & 90.60$\pm$0.48  \\
\rowcolor{LightCyan}  protoLP (ours)               & WRN-28-10      & \textcolor{red}{87.69$\pm$0.23} & \textcolor{red}{90.82$\pm$0.15}\\
\bottomrule
\end{tabular}
}
\label{tab:cifar}
\end{minipage}
\end{table}

\vspace{0.1cm}
\noindent\textbf{Semi-supervised Learning.} In this setting, one has an access to an additional set of unlabeled samples along with each test task. 
These unlabeled samples may  contain both the target task category or other categories. 
Table \ref{tab:result2} summarizes the performance of our methods and SOTA semi-supervised FSL methods, and shows  
%
that protoLP outperforms such baselines in all settings by a large margin (ResNet-12 backbone). The gain varies between 3\% and 6\% on mini-ImageNet 1-shot protocol due to  
%
capturing data manifold by using learnable graph with extra unlabeled samples. 
On WRN-28-10, protoLP also outperforms other methods by a fair margin in the 1-shot setting, \eg, between 1.3\% and 3.5\% on mini-ImageNet 1-shot.  
PT+MAP~\cite{hu2020leveraging} offers no results on semi-supervised learning so we use  iLPC~\cite{lazarou2021iterative} that provides the code for PT+MAP (WRN-28-10) in that setting.
On tiered-ImageNet, the larger number of categories resulted in randomly chosen diverse unlabeled samples which  had negative effect on  support/query sets. 

\begin{table*}[htbp]
  \centering
  \caption{Comparison of test accuracy against state-of-the-art methods for 1-shot and 5-shot classification under the semi-supervised few-shot learning setting. CUB 5-shot omitted: no class has the required 70 examples.}
  \vspace{-0.1cm}
  \resizebox{1\textwidth}{!}{
    \begin{tabular}{r r r ll cc cc cc}
    \toprule
          &      &       & \multicolumn{2}{c}{mini-ImageNet} & \multicolumn{2}{c}{tiered-ImageNet} & \multicolumn{2}{c}{CIFAR-FS} & \multicolumn{2}{c}{CUB}\\
    \midrule
    Methods  &Backbone & Setting & 1-shot &  5-shot & 1-shot &  5-shot & 1-shot &  5-shot & 1-shot &  5-shot\\
    \midrule
LR+ICI~\cite{wang2020instance}  & ResNet-12 & 30/50 & $67.57_{\pm0.97}$ & $79.07_{\pm0.56}$ & $83.32_{\pm0.87}$ & $89.06_{\pm0.51}$ & $75.99_{\pm0.98}$ & $84.01_{\pm0.62}$ & $88.50_{\pm0.71}$ & -      \\
iLPC~\cite{lazarou2021iterative}      & ResNet-12 & 30/50 & $70.99_{\pm0.91}$ & $81.06_{\pm0.49}$ & $85.04_{\pm0.79}$ & $89.63_{\pm0.47}$ & $78.57_{\pm0.80}$ & $85.84_{\pm0.56}$ & $90.11_{\pm0.64}$ & -      \\
\rowcolor{LightCyan}  protoLP (ours)       & ResNet-12 & 30/50 & $\textcolor{red}{72.21}_{\pm0.88}$ & $\textcolor{red}{81.48}_{\pm0.49}$ & $\textcolor{red}{85.22}_{\pm0.79}$ & $\textcolor{red}{89.64}_{\pm0.46}$ & $\textcolor{red}{80.02}_{\pm0.88}$ & $\textcolor{red}{86.16}_{\pm0.53}$ & $\textcolor{red}{90.26}_{\pm0.65}$ & -      \\
\midrule
LR+ICI~\cite{wang2020instance}  & WRN-28-10  & 30/50 & $81.31_{\pm0.84}$ & $88.53_{\pm0.43}$ & $88.48_{\pm0.67}$ & $92.03_{\pm0.43}$ & $86.03_{\pm0.77}$ & $89.57_{\pm0.53}$ & $90.82_{\pm0.59}$ & -      \\
PT+MAP~\cite{hu2020leveraging} & WRN-28-10  & 30/50 & $83.14_{\pm0.72}$ & $88.95_{\pm0.38}$ & $89.16_{\pm0.61}$ & $92.30_{\pm0.39}$ & $87.05_{\pm0.69}$ & $89.98_{\pm0.49}$ & $91.52_{\pm0.53}$ & -      \\
iLPC~\cite{lazarou2021iterative}     & WRN-28-10  & 30/50 & $83.58_{\pm0.79}$ & $\textcolor{red}{89.68}_{\pm0.37}$ & $89.35_{\pm0.68}$ & $\textcolor{red}{92.61}_{\pm0.39}$ & $87.03_{\pm0.72}$ & $90.34_{\pm0.50}$ & $91.69_{\pm0.55}$ & -     \\
\rowcolor{LightCyan}  protoLP (ours)     & WRN-28-10  & 30/50 & $\textcolor{red}{84.25}_{\pm0.75}$ & $89.48_{\pm0.39}$ & $\textcolor{red}{90.10}_{\pm0.63}$ & $92.49_{\pm0.40}$ & $\textcolor{red}{87.92}_{\pm0.69}$ & $\textcolor{red}{90.51}_{\pm0.48}$ & $\textcolor{red}{92.01}_{\pm0.57}$ & -     \\
\bottomrule
    \end{tabular}
    }
    \label{tab:result2}
\end{table*}

\begin{table}[htbp]
  \centering
  \caption{Comparison of test accuracy against state-of-the-art methods (DenseNet and MobileNet, 1- and 5-shot protocols). Notice SimpleShot is an inductive method based on the above backbone.
  }
  \vspace{-0.1cm}
  \resizebox{0.49\textwidth}{!}{
    \begin{tabular}{r ll cc}
    \toprule
          & \multicolumn{2}{c}{mini-ImageNet} & \multicolumn{2}{c}{tiered-ImageNet} \\
    \midrule
    Methods (DenseNet)  & 1-shot &  5-shot & 1-shot &  5-shot \\
    \midrule
    SimpleShot~\cite{wang2019simpleshot} & 65.77 $\pm$ 0.19  &82.23 $\pm$ 0.13 &71.20 $\pm$ 0.22  &86.33 $\pm$ 0.15 \\
    LaplacianShot~\cite{ziko2020laplacian} & 75.57 $\pm$ 0.19  &84.72 $\pm$ 0.13 &80.30 $\pm$ 0.20 &87.93 $\pm$ 0.15\\
    RAP-LaplacianShot~\cite{hong2021reinforced}	&	75.58 $\pm$ 0.20	& 85.63 $\pm$ 0.13 & - &	-\\
    TAFSSL(PCA)~\cite{lichtenstein2020tafssl} &70.53 $\pm$ 0.25 &80.71 $\pm$ 0.16 &80.07 $\pm$ 0.25 &86.42 $\pm$ 0.17\\
    TAFSSL(ICA)~\cite{lichtenstein2020tafssl}  &72.10 $\pm$ 0.25 &81.85 $\pm$ 0.16 &80.82 $\pm$ 0.25 &86.97 $\pm$ 0.17\\
    $\!\!\!$TAFSSL(ICA+MSP)~\cite{lichtenstein2020tafssl} & 77.06 $\pm$ 0.26 & 84.99 $\pm$ 0.14 & 84.29 $\pm$ 0.25 & 89.31 $\pm$ 0.15\\
    \rowcolor{LightCyan} protoLP (ours)  & \textcolor{red}{79.27 $\pm$ 0.27} & \textcolor{red}{85.88 $\pm$ 0.14} & \textcolor{red}{86.17 $\pm$ 0.25} & \textcolor{red}{90.50 $\pm$ 0.15}\\
    \hline
    \hline
    Methods (MobileNet)  & 1-shot &  5-shot & 1-shot &  5-shot \\
    \midrule
SimpleShot~\cite{mishra2017simple} &61.55 $\pm$ 0.20 &77.70 $\pm$ 0.15 &69.50 $\pm$ 0.22 &84.91 $\pm$ 0.15 \\
LaplacianShot~\cite{ziko2020laplacian} &70.27 $\pm$ 0.19& 80.10 $\pm$ 0.15 &79.13 $\pm$ 0.21 &86.75 $\pm$ 0.15 \\
\rowcolor{LightCyan} protoLP (ours)  &72.04 $\pm$ 0.23 &82.11 $\pm$ 0.20 &80.68 $\pm$ 0.24 &87.45 $\pm$ 0.19 \\
    \bottomrule
    \end{tabular}
    }
    \label{tab:result3}
\end{table}

\subsection{Ablation Studies}

\subsubsection{Uniform Class Prior} 

As many methods use Optimal Transport (OT)  to leverage the uniform prior on the class distribution, we demonstrate how these methods benefit from the prior by Sinkhorn distance. 
To further investigate the potential of protoLP, we conduct ablations on mini-ImageNet 
to compare  FSL with Sinkhorn (uniform class prior) \vs no Sinkhorn (no prior).  Tables~\ref{tab:AblationStudy} and \ref{tab:AblationStudy2} show  
that OT  improves results especially in 1-shot classification when the features are not discriminative enough (ResNet-12). For example, OT improves  performances of EASE by 13\% and iLPC by 4.5\% in mini-ImageNet. Notice that OT only  boost  protoLP by  0.7\% in mini-ImageNet. In 5-shot setting, the performance gains from OT are reduced but they follow the same pattern as in 1-shot setting. As  WRN-28-10 backbone  yields good performance, gains from OT are lesser than for ResNet-12.

The performance gain of OT on protoLP is small while overall results of protoLP are high. Section \ref{sec:unb} shows that results of many OT-based methods degrade significantly when the uniform class prior is used and the real class distribution does not follow it. Our protoLP is an exception. 

\subsubsection{Comparisons with the classical LP}
Our protoLP improves results significantly compared to classical LP (no prototypes used) in Table \ref{tab:AblationStudy}. On ResNet-12, protoLP gains 9\% and 4\% over LP on mini-ImageNet (1- and 5-shot prot.) On WRN-28-10, protoLP gains 7.5\% and 3.7\% on mini-ImageNet (1- and 5-shot prot.) 


\subsubsection{Data Augmentation for Inference}
\label{sec:da}
Some methods apply data augmentation techniques to boost inference. In \tabref{result}, we report results of protoLP$^*$ with data augmentation. The protoLP$^*$ and EASY$^*$ use random resized crops from each image. We
obtain multiple versions of each feature vector and average them. MCT  augments both the input image and the intermediate model features. Based on these augmentations, MCT learn a meta-learning confidence with input-adaptive distance metric. ODC employs spatial pyramid pooling to augment  intermediate features of the backbones. The use of augmentation (from data or from models) in the inference stage improves performance. \tabref{result} shows this effect is particularly evident in the 1-shot classification of mini-ImageNet where  protoLP$^*$ outperforms the protoLP by nearly 14\%.
%
\begin{table}
\caption{The uniform class prior (Sinkhorn \vs no Sinkhorn).}
\vspace{-0.1cm}
\resizebox{0.49\textwidth}{!}{
\begin{tabular}{r c l l l}
\toprule
& &&\multicolumn{2}{l}{$\quad\;$mini-ImageNet}                                               \\
\midrule
Method                     & Sinkhorn & Backbone & 1-shot        & 5-shot        \\
\midrule
LP & & ResNet-12 & 61.09$\pm$0.70 & 75.32$\pm$0.50 \\
EASE & & ResNet-12 & 57.00$\pm$0.26 & 75.07$\pm$0.21 \\
EASE & \checkmark& ResNet-12 & 70.47$\pm$0.30 & 80.73$\pm$0.16 \\
iLPC & & ResNet-12 & 65.57$\pm$0.89 & 78.03$\pm$0.54\\
iLPC & \checkmark & ResNet-12 & 69.79$\pm$0.99 & 79.82$\pm$0.55 \\
\rowcolor{LightCyan} protoLP &             & ResNet-12      & \textcolor{red}{70.04$\pm$0.29} & \textcolor{red}{79.80$\pm$0.16} \\
\rowcolor{LightCyan} protoLP  & \checkmark & ResNet-12 & \textcolor{red}{70.77$\pm$0.30} & \textcolor{red}{80.85$\pm$0.16} \\
\midrule
LP & & WRN-28-10 & 74.24$\pm$0.68 & 84.09$\pm$0.42 \\
PT-MAP & &WRN-28-10 & 82.92$\pm$0.26	& 88.82$\pm$0.13\\
EASE & & WRN-28-10 & 67.42$\pm$0.27 & 84.45$\pm$0.18 \\
EASE &\checkmark & WRN-28-10 & 83.00$\pm$0.21 & 88.92$\pm$0.13 \\
iLPC &                & WRN-28-10      & 78.29$\pm$0.76 &87.62$\pm$0.41  \\
iLPC & \checkmark & WRN-28-10 & 83.05$\pm$0.79 & 88.82$\pm$0.42\\
\rowcolor{LightCyan} protoLP &              & WRN-28-10      & \textcolor{red}{81.91$\pm$0.25} & \textcolor{red}{87.85$\pm$0.13}\\
\rowcolor{LightCyan} protoLP & \checkmark           & WRN-28-10      & \textcolor{red}{83.07$\pm$0.25} & \textcolor{red}{89.04$\pm$0.13}\\
\bottomrule
\end{tabular}
}
\label{tab:AblationStudy}
\end{table}

\begin{table}
\caption{The uniform class prior (Sinkhorn \vs no Sinkhorn).}
\vspace{-0.1cm}
\resizebox{0.49\textwidth}{!}{
\begin{tabular}{r c l l l}
\toprule
& &&\multicolumn{2}{l}{$\quad\;$tiered-ImageNet}                                               \\
\midrule
Method                     & Sinkhorn & Backbone & 1-shot        & 5-shot        \\
\midrule
LP & & ResNet-12 & 73.29$\pm$0.35 & 86.32$\pm$0.30 \\
EASE & & ResNet-12 & 69.74$\pm$0.31 & 85.17$\pm$0.21 \\
EASE & \checkmark& ResNet-12 & 84.54$\pm$0.27 & 89.63$\pm$0.15 \\
\rowcolor{LightCyan} protoLP &             & ResNet-12      & \textcolor{red}{83.59$\pm$0.25} & \textcolor{red}{88.60$\pm$0.15} \\ 
\rowcolor{LightCyan} protoLP  & \checkmark & ResNet-12 & \textcolor{red}{84.69$\pm$0.29} & \textcolor{red}{89.47$\pm$0.15} \\
\midrule
LP & & WRN-28-10 & 76.24$\pm$0.30 & 85.09$\pm$0.25 \\
EASE & & WRN-28-10 & 75.87$\pm$0.29 & 85.17$\pm$0.21 \\
EASE &\checkmark & WRN-28-10 & 88.96$\pm$0.23 & 92.63$\pm$0.13 \\
\rowcolor{LightCyan} protoLP &              & WRN-28-10      & \textcolor{red}{87.91$\pm$0.25} & \textcolor{red}{91.60$\pm$0.13}\\
\rowcolor{LightCyan} protoLP & \checkmark           & WRN-28-10      & \textcolor{red}{89.04$\pm$0.23} & \textcolor{red}{92.80$\pm$0.13}\\
\bottomrule
\end{tabular}
}
\label{tab:AblationStudy2}
\end{table}

\begin{table}[htbp]
  \centering
  \caption{Test accuracy against the state of the art in the class-unbalanced setting  (WRN-28-10, 1- and 5-shot protocols).}
   \vspace{-0.1cm}
  \resizebox{0.40\textwidth}{!}{
    \begin{tabular}{r ll cc}
    \toprule
          & \multicolumn{2}{c}{mini-ImageNet} & \multicolumn{2}{c}{tiered-ImageNet} \\
    \midrule
    Methods  & 1-shot &  5-shot & 1-shot &  5-shot \\
    \midrule
Entropy-min & 60.4  &76.2  &62.9  &77.3 \\
PT-MAP  &60.6 & 66.8  &65.1  &71.0 \\
LaplacianShot  &68.1 & 83.2 &73.5  &86.8 \\
TIM & 69.8  &81.6  &75.8 & 85.4 \\
BD-CSPN  &70.4  &82.3 &75.4 & 85.9 \\
$\alpha$-TIM & 69.8 & 84.8 & 76.0 & 87.8\\
\rowcolor{LightCyan} protoLP (ours)  & \textcolor{red}{73.7}& \textcolor{red}{85.2} & \textcolor{red}{81.0} & \textcolor{red}{89.0}\\
\bottomrule
    \end{tabular}
    }
    \label{tab:unbalance}
\end{table}

\subsubsection{Evaluations on Class-unbalanced Setting}
\label{sec:unb}
Below, we follow the same unbalanced setting as \cite{veilleux2021realistic} where the query set is randomly distributed, following a Dirichlet distribution parameterized by $\alpha=2$. The performance 
is evaluated by computing the average accuracy over 10,000 few-shot tasks. Table~\ref{tab:unbalance} shows that due to the use of the uniform prior on the class distribution,  PT-MAP~\cite{hu2020leveraging} looses 18\% accuracy maximum in the unbalanced setting. Other methods also loose few percents on mini-ImageNet, tiered-ImageNet and CUB with the WRN-28-10 backbone. 
Our protoLP outperforms other models by 3.3\%, 5.6\% and 6.5\% on 1-shot protocol in reported datasets.

\begin{table}
\centering
\caption{Test accuracy against the state of the art in the class unbalanced setting (ResNet-12, 1-shot protocols, CUB).}
\resizebox{0.40\textwidth}{!}{
\begin{tabular}{rcc}
\hline CUB & \multicolumn{1}{c}{ unbalanced } & \multicolumn{1}{c}{ balanced } \\
\cline { 2 - 3} Method & 1-shot  & 1-shot  \\
\hline PT-MAP~\cite{hu2020leveraging} & $65.1$ & $85.5$ \\
LaplacianShot~\cite{ziko2020laplacian} & $73.7$ & $78.9$  \\
BD-CSPN~\cite{liu2020prototype} & $74.5$ & $77.9$  \\
TIM~\cite{boudiaf2020information} & $74.8$ & $80.3$  \\
$\alpha$-TIM~\cite{veilleux2021realistic} & $75.7$ & $-$\\
\hline
\rowcolor{LightCyan} protoLP & $\textcolor{red}{82.22}$  & $\textcolor{red}{90.13}$ \\
\hline
\end{tabular}
}
\label{tab:unbalance2}
\end{table}

\subsubsection{DenseNet/MobileNet (Multi-class Pre-training)}
Compared with other transductive methods based on backbones with the meta-learning framework, TAFSSL~\cite{lichtenstein2020tafssl} uses SimpleShot~\cite{wang2019simpleshot} backbones, 
and so we also extract features by backbones (DenseNet, MobileNet) from SimpleShot, 
which directly train backbone with a nearest-neighbor classifier instead of meta-learning (as  ResNet-12, ResNet-18, WRN-28-10 from S2M2-R~\cite{mangla2020charting}). 
Thus, below we show that protoLP is independent of the way backbones are trained. 
Table~\ref{tab:result3} shows that  protoLP is superior to counterparts with prototypes (TAFFSL) and label propagation (LaplacianShot) in all settings, especially in 1-shot protocol, outperforming them by a large margin (DenseNet backbone), \eg,  3.7\% and 2.2\% on mini-ImageNet 1-shot protocol, and  5.8\% and 1.8\% in tiered-ImageNet 1-shot protocol.


\subsection{Inference Time and Convergence}
\begin{figure}[t]
     \centering
     \includegraphics[width=0.40\textwidth]{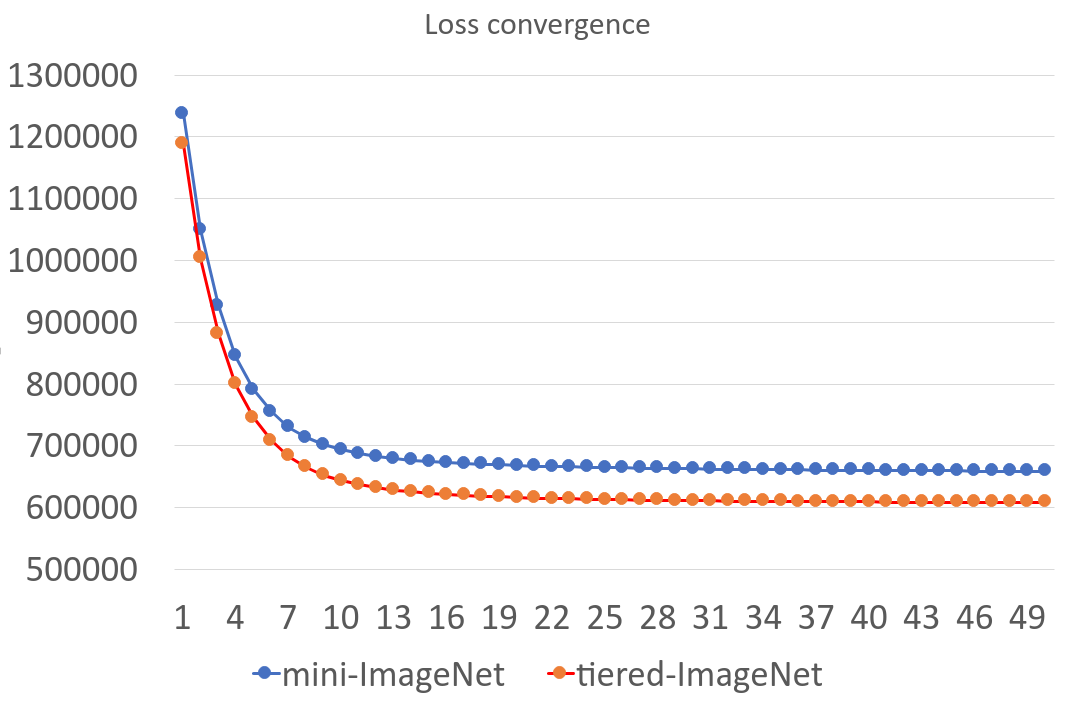}
      \caption{The loss curve (mini-ImageNet, tiered-ImageNet).}
     \label{fig:convergence}
\end{figure}

The computational complexity of protoLP depends only on the feature dimension and the number of samples. Thus, for different datasets, the computational cost appears equal.  
Our protoLP takes only a few of milliseconds (about 10$\times$ faster than iLPC and ICI, as shown in the supplementary material, which does not impose any burden on typical applications. 
%
%
Finally, Fig.~\ref{fig:convergence} shows the value of loss in Eq.~\eqref{eq:protoLP} \wrt the iteration number. The loss converges fast. 

  
\section{Conclusions}
%
In this paper, we have pointed out disadvantages of prototype-based methods and label propagation based methods for transductive FSL.
To overcome these drawbacks, we have presented a unified framework combining the prototype-based methods and label propagation based methods within a single objective.
Our protoLP  inherits  advantages of individual prototype refinement and  label propagation steps while avoiding the disadvantages of the bias in estimation of prototypes and  the fixed graph bias. 
Our protoLP performs well also under non-uniform class priors unlike Sinhorn-based methods. The protoLP works with  different backbones and is a plug-and-play module for the inference step of FSL. 

\paragraph{Acknowledgements.} 
HZ is supported by an Australian Government Research Training Program (RTP) Scholarship. 
PK is in part funded by CSIRO's Machine Learning and Artificial Intelligence Future Science Platform (MLAI FSP) Spatiotemporal Activity.

{\small
\bibliographystyle{ieee_fullname}
\bibliography{egbib}
}

\clearpage


%
\title{Transductive Few-shot Learning with Prototype-based Label Propagation by Iterative Graph Refinement (Supplementary Material)}

\author{
{Hao Zhu$^{\dagger, \S}$, $\quad$Piotr Koniusz$^{*, \S,\dagger}$}\\
{$^{\dagger}$Australian National University $\quad^\S$Data61\heart CSIRO}\\
allenhaozhu@gmail.com, $\;$firstname.lastname@anu.edu.au\\
}

\maketitle

\setcounter{table}{9}
\setcounter{equation}{18}
\setcounter{section}{5}

%
Below are additional details regarding our model and experiments.

\section{Datasets}
\paragraph{mini-ImageNet.} This dataset \citelatex{vinyals2016matching_supp,ravi2017optimization_supp} is widely used in few-shot classification \citelatex{maxexp_supp,simon2022noise_supp,yao_cvpr2022_supp,simon2022wacv_supp,hong_tmm_supp,zhang2022accv_supp,simon_cvpr2023_supp}. It contains 100 randomly chosen classes from ImageNet~\citelatex{russakovsky2015imagenet_supp}. There are 64 training (base) classes, 16 validation (novel) classes, and 20 test (novel) classes among the 100 classes. There are 600 images in each class. We adopt the split provided in \citelatex{ravi2017optimization_supp}.

\vspace{-0.4cm}
\paragraph{tiered-ImageNet.}  ImageNet  with a hierarchical structure was used to create the tiered-ImageNet. Categories of classes are divided into 34 categories, each of which contains 351, 97, and 160 classes for training, validation, and testing, respectively. Please note the training and test classes  are semantically disjoint. We follow the common split in \citelatex{chen2021eckpn_supp} and 84 by 84px resolution.

\vspace{-0.4cm}
\paragraph{CIFAR-FS.} This dataset
has 100 classes, each with 600 examples in CIFAR-100~\citelatex{krizhevsky2009learning_supp}, on which this dataset is based on. We use the 64 training, 16 validation, and 20 test classes provided by \citelatex{chen2019closer_supp}. 

\vspace{-0.4cm}
\paragraph{CUB.} There are 200 classes, each representing a bird species, in this fine-grained dataset. Following the setting in \citelatex{chen2019closer_supp} , we divide our classes into three groups: 100 training, 50 validation, and 50 testing classes. 

\section{Feature Extraction and Pre-processing}
\paragraph{ResNet-12A} $\!\!\!$is the pre-trained backbone network used in \citelatex{wang2020instance_supp}. For all of our transductive and semi-supervised experiments using this network, we adopt exactly the same pre-processing as \citelatex{wang2020instance_supp}, which includes normalizing feature vectors by their  $\ell_{2}$ norms.

\vspace{-0.4cm}
\paragraph{WRN-28-10} $\!\!\!$is the pre-trained network used in \citelatex{mangla2020charting_supp} and \citelatex{hu2020leveraging_supp}. To provide fair comparisons with PT+MAP~\citelatex{hu2020leveraging_supp}, we adopt exactly the same pre-processing as \citelatex{hu2020leveraging_supp}. In the all experiments, we apply the power transform and normalize feature vectors by their  $\ell_{2}$ norms.

\vspace{-0.4cm}
\paragraph{DenseNet} $\!\!\!$is the pre-trained network used in \citelatex{lichtenstein2020tafssl_supp} and \citelatex{wang2019simpleshot_supp}. To provide fair comparisons with TASFFL~\citelatex{lichtenstein2020tafssl_supp}, we adopt exactly the same pre-processing. In the all experiments, we decentre feature vectors by using the center in the training set and normalize decentred feature vectors by their $\ell_{2}$ norms.

\paragraph{MobilieNet} is the pre-trained network used in \citelatex{wang2019simpleshot_supp}. To provide fair comparison with LaplacianShot~\citelatex{ziko2020laplacian_supp}, we adopt exactly the same pre-processing. In the all experiments, we follow the setting as same as other backbones.

\section{Hyper-parameters}
In Eq.~\eqref{eq:AGSSL},  parameter $\lambda$ is in charge of regularizing the graph. For the balanced class setting, we simply set $\lambda=1$. For the unbalanced class setting, we set $\lambda=0.5$. 

Hyper-parameter $\alpha$ is used to control updating prototypes $\mC$ in Eq.~\eqref{eq:abc}. We empirically found that $\alpha$ basically does not impact the final result but the convergence speed. In this paper, we set it as 0.2 for all experiments.

\section{Note on Fair Comparisons}

During our experimental studies, we noticed the importance of fair comparisons by ensuring the common testbed. 
Below we talk about common issues.
\begin{itemize}
    \item \textbf{Some comparisons use different networks}. In some papers, ResNet-18 is  compared directly with other methods using ResNet-12. In Table \ref{tab:result} (our main paper), we show that ResNet-18 has some advantages over ResNet-12 in some cases.
    \item \textbf{Methods are mainly compared under the class-balanced prior}. For ease of understanding, transductive FSL evolves directly from the setting of inductive FSL. Class-balanced queries are not an issue for inductive FSL because these methods treat queries one-by-one and ignore the class distribution of queries. However, the same setting in transductive setting introduces class-balanced prior. Many approaches over the last few years have focused on how to exploit this prior. We argue that this prior is unreasonable for transductive FSL, since it is rare for queries to follow a uniform class distribution. We can see the power of optimal transport under this prior i numerous works~\citelatex{zhu2022ease_supp,hu2020leveraging_supp,lazarou2021iterative_supp}. We can also see that this technique has the negative effect for queries that do not conform to the uniform class prior (\eg, PT-MAP~\citelatex{hu2020leveraging_supp} drop 17\% in mini-ImageNet under the unbalanced setting). Without the optimal transport, these methods also loose performance in the balanced setting.
\end{itemize}

We hope that bringing attention to these evaluation issues will help researchers avoid following the unrealistic settings and move toward fairer evaluation protocols and models. We encourage the community to compare different methods with under the same testbed.

\section{The motivation of Parameterized Label Prediction}
From a theoretical point of view, if we do not use parameterized label prediction, we need to optimize:
%
\begin{equation}
\min_{\widetilde{\mY}} \frac{1}{2}\lVert\widetilde{\mY}_L-\mY_L\rVert_F^2+\frac{\lambda}{2} \Tr(\widetilde{\mY}^{\top}\left(\boldsymbol{I}-\lambda\boldsymbol{Z} \boldsymbol{\Lambda}^{-1} \mZ^\top\right)\widetilde{\mY}).
\end{equation}

\noindent
The closed form solution is $\left(\boldsymbol{I}-\lambda\mZ \boldsymbol{\Lambda}^{-1} \mZ^\top\right)^{-1}\mY$. The computational complexity of the inverse matrix is $n^3$. and the $rank(\mZ \boldsymbol{\Lambda}^{-1} \mZ^\top)\!=\!5$ (for 5-way problem). 
Thus we can use linear function $\mA$ to do propagation on $\mY$ and reduce computational complexity from $n^3$ to $c^3$ where $n\!\gg\!c$. Projection to low-dimensional space limits the number of unnecessary parameters.
%
%
%
Experiments in Table \ref{tab:overfit} confirm the parameterized LP works better.
\begin{table}[htbp]
\centering
\caption{Test accuracy on standard LP and parameterized LP setting (ResNet-12 backbone).}
  \resizebox{0.47\textwidth}{!}{
      \renewcommand{\arraystretch}{0.5}
    \begin{tabular}{r ll ll ll}
    \toprule
          & \multicolumn{2}{c}{mini-ImageNet} & \multicolumn{2}{c}{tiered-ImageNet}   & \multicolumn{2}{c}{CUB} \\
    \midrule
    Methods (ResNet-18) & 1-shot & 5-shot & 1-shot &  5-shot& 1-shot &  5-shot \\
    \midrule
    Ours (non-paramerized) &69.41 &79.52 &83.60 &88.56 &88.50 &91.70 \\
    \rowcolor{LightCyan} Ours (parameterized) &70.04 & 79.80 & 84.04 & 88.72 & 88.85 & 91.93\\
    \bottomrule
    \end{tabular}
    }
    \label{tab:overfit}
\end{table}

\section{Inference Time}
Table~\ref{tab:time} provides  the average inference time (1$\times$ AMD 3600 CPU) for the 1- and 5-shot tasks on mini-ImageNet (ResNet-12 and WRN-28-10).
\begin{table}[htbp]
\centering
\caption{Average inference time (in seconds) for the 1-shot and 5-shot tasks in mini-ImageNet dataset with different backbones.}
\hspace{0.3cm}
  \resizebox{0.35\textwidth}{!}{
\begin{tabular}{r c c c c}
\toprule
Backbone & \multicolumn{2}{c }{ResNet-12} & \multicolumn{2}{c}{WRN-28-10} \\
\midrule
Shot     & 1             & 5            & 1             & 5            \\
\hline
iLPC~\citelatex{lazarou2021iterative_supp}     & 4.5e-2        & 5.6e-2       & 5.5e-2        & 7.0e-2\\
ICI~\citelatex{wang2020instance_supp}     & 3.4e-2        & 4.2e-2       & 4.1e-2        & 5.2e-2\\
\rowcolor{LightCyan} protoLP     & 4.7e-3        & 5.8e-3       & 6.2e-3        & 6.8e-3 \\ 
\bottomrule
\end{tabular}
}
\label{tab:time}
\vspace{-0.2cm}
\end{table}

{\small
\bibliographystylelatex{ieee_fullname}
\bibliographylatex{latex}
}

\end{document}